\documentclass[lettersize,journal]{IEEEtran}
\usepackage{amsmath,amsfonts}
\usepackage{algorithmic}
\usepackage{algorithm}
\usepackage{array}
\usepackage[caption=false,font=normalsize,labelfont=sf,textfont=sf]{subfig}
\usepackage{textcomp}
\usepackage{stfloats}
\usepackage{url}
\usepackage{verbatim}
\usepackage{graphicx}
\usepackage{cite}
\usepackage{multirow}

\begin{document}

\title{Graph-Based Interaction-Aware Multimodal 2D Vehicle Trajectory Prediction using Diffusion Graph Convolutional Networks}

\author{Keshu Wu, Yang Zhou, Haotian Shi*, Xiaopeng Li, and Bin Ran
\thanks{*Corresponding author: Haotian Shi (Email: hshi84@wisc.edu)}
\thanks{Keshu Wu, Haotian Shi, Xiaopeng Li, Bin Ran are with the Department of Civil and Environmental Engineering, University of Wisconsin-Madison, WI, USA.}
\thanks{Yang Zhou is with the Zachry Department of Civil and Environmental Engineering, Texas A$\&$M University, TX, USA.}}

\markboth{IEEE Transactions on Intelligent Vehicles, manuscript}%
{Shell \MakeLowercase{\textit{et al.}}: A Sample Article Using IEEEtran.cls for IEEE Journals}


\maketitle

\begin{abstract}
Predicting vehicle trajectories is crucial for ensuring automated vehicle operation efficiency and safety, particularly on congested multi-lane highways. In such dynamic environments, a vehicle's motion is determined by its historical behaviors as well as interactions with surrounding vehicles. These intricate interactions arise from unpredictable motion patterns, leading to a wide range of driving behaviors that warrant in-depth investigation. This study presents the Graph-based Interaction-aware Multi-modal Trajectory Prediction (GIMTP) framework, designed to probabilistically predict future vehicle trajectories by effectively capturing these interactions. Within this framework, vehicles' motions are conceptualized as nodes in a time-varying graph, and the traffic interactions are represented by a dynamic adjacency matrix. To holistically capture both spatial and temporal dependencies embedded in this dynamic adjacency matrix, the methodology incorporates the Diffusion Graph Convolutional Network (DGCN), thereby providing a graph embedding of both historical states and future states. Furthermore, we employ a driving intention-specific feature fusion, enabling the adaptive integration of historical and future embeddings for enhanced intention recognition and trajectory prediction. This model gives two-dimensional predictions for each mode of longitudinal and lateral driving behaviors and offers probabilistic future paths with corresponding probabilities, addressing the challenges of complex vehicle interactions and multi-modality of driving behaviors. Validation using real-world trajectory datasets demonstrates the efficiency and potential.
\end{abstract}


\section{Introduction}
\IEEEPARstart{V}{ehicle} trajectory prediction, as a critical aspect of intelligent transportation systems, is imperative for enhancing traffic safety, reducing congestion, and promoting sustainable transportation. To enable the efficient operation of vehicles, it is essential to have a comprehensive understanding of the driving environment, as well as the ability to generate precise predictions of the movements \cite{ref1,refa1} of surrounding objects. This knowledge is vital for intelligent decision-making, trajectory planning, and control during CAV operation, and will ultimately contribute to the development of an intelligent and dependable transportation network \cite{refa2,refa3}. 

Recognizing the intricacies of the driving environment underscores the fact that a vehicle's trajectory is shaped by multifaceted influences. It is evident that the movement of a vehicle is not only influenced by its historical trajectory but also by the motion states and maneuvers of neighboring vehicles, particularly in dense traffic conditions \cite{ref2, ref3}. Nevertheless, navigating through complex traffic environments with multiple non-cooperative vehicles creates inherent uncertainty due to the numerous latent variables involved \cite{ref4}. Accurately predicting the future trajectory of a target vehicle has gained significant attention and is becoming a critical aspect for enhancing traffic safety \cite{ref5, ref6}. The complex nature of automated driving and the dynamic and dense traffic environments pose challenges for current trajectory prediction research, particularly in the following two aspects: (i) interaction between vehicles \cite{ref7}, (ii) multi-modality of driving behaviors \cite{ref8}. 

\begin{figure}[!t]
\centering
\includegraphics[scale=0.35]{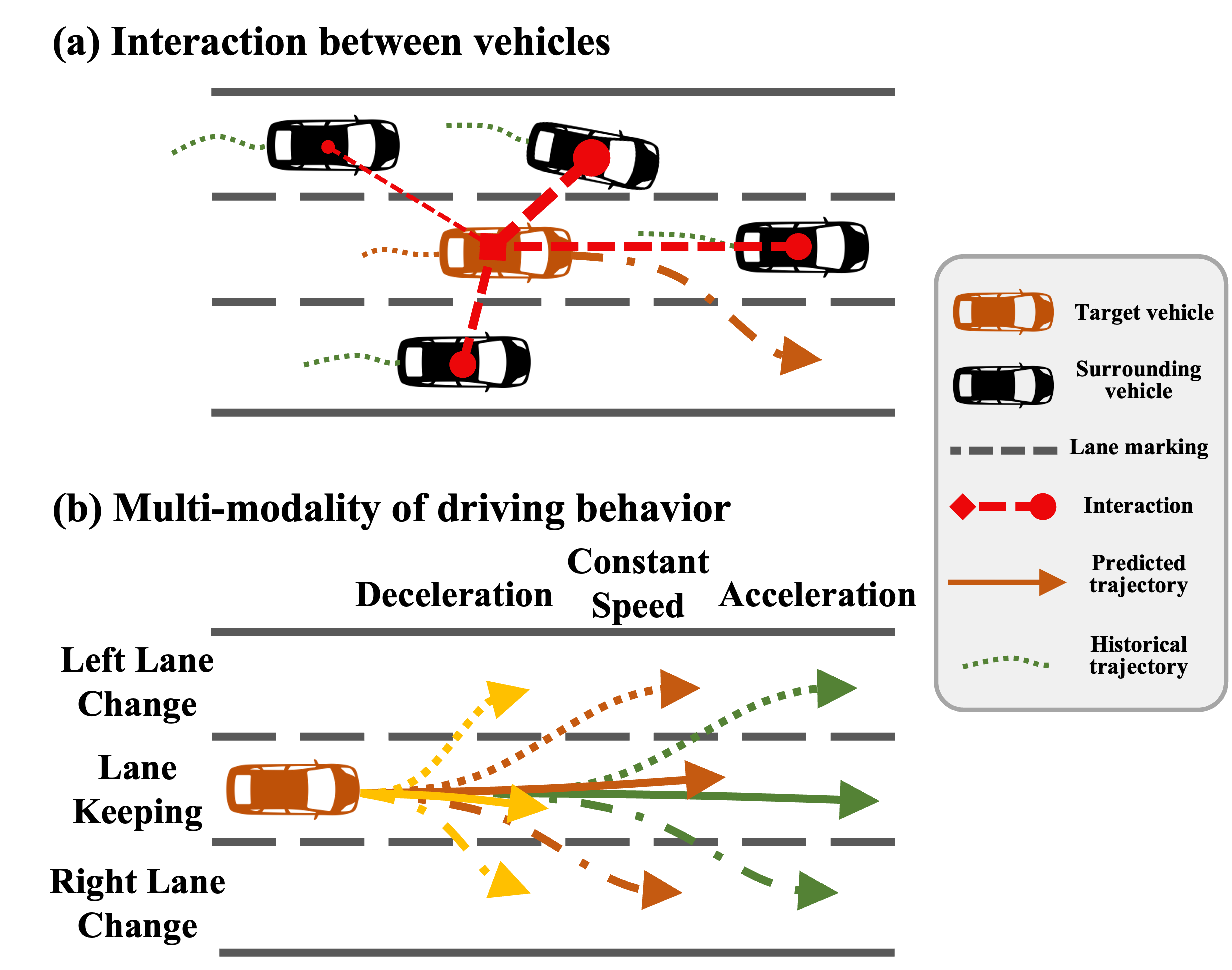}
\caption{Vehicle interaction and behavior multimodality.}
\label{fig_1}
\end{figure}

\begin{figure}[!h]
\centering
\includegraphics[scale=0.65]{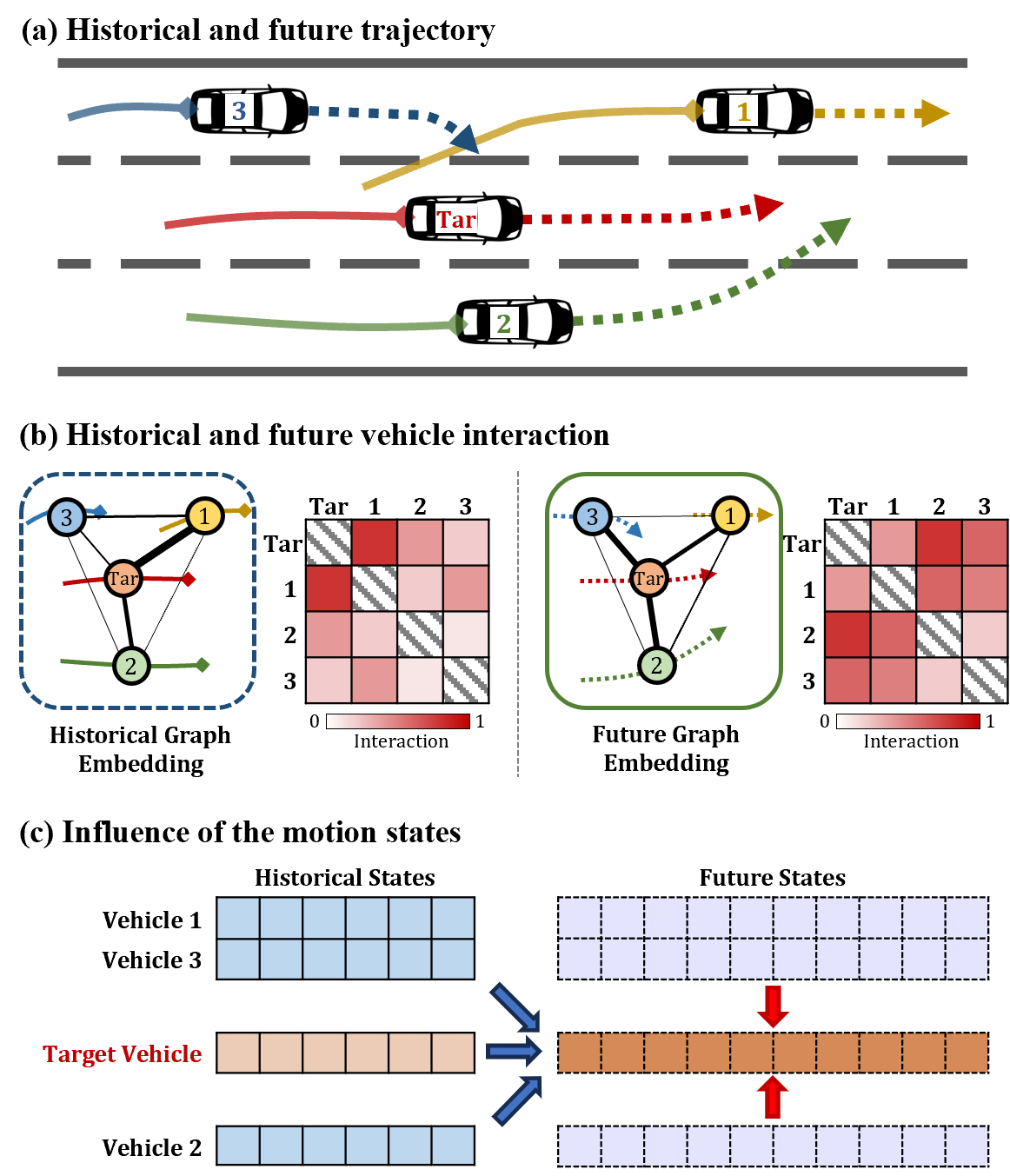}
\caption{Key concepts of this paper. }
\label{fig_2}
\end{figure}

\textbf{Interaction.} Considering the interaction between agents helps to predict a socially aware trajectory \cite{ref7, ref9}. Traditional prediction methods, such as physics-based models \cite{ref10, ref11} and Kalman filter method \cite{ref12}, alongside classic machine learning-based approaches \cite{ref13, ref14, ref15, ref16}, are inadequate in handling complex prediction contexts \cite{ref6}. Deep learning has recently emerged as a preferred tool for trajectory prediction in complex contexts, thanks to its proficiency in learning intricate features, incorporating factors related to physics, road geometry, and interactions between vehicles. To effectively utilize the input historical information, transformation into representations that underscore temporal and spatial correlations is essential. Representations such as time series sequence \cite{ref17, ref18}, occupancy girds \cite{ref19, ref20}, and rasterized image \cite{ref21, ref22} have been previously utilized. Recurrent neural networks (RNN) including its variants Long Short-Term Memory (LSTM) and Gated Recurrent Unit (GRU) are particularly suitable for capturing temporal correlations in sequential data. To capture the spatial correlation, including inter-vehicle interaction or vehicle-road interaction, approaches including Convolutional Neural Network (CNN) based methods \cite{ref7, ref19, ref30, ref31, ref32}, attention-driven methods \cite{ref23, ref24, ref25, ref33}, and Graph Neural Network (GNN) based methods \cite{ref26, ref27, ref28, ref29} are proposed. Particularly, GNN-based methods show promise in deciphering non-Euclidean spatial dependencies, rendering them suitable for simulating interactions. A shortcoming of most graph-based approaches is their reliance on adjacency matrices, constructed solely based on neighborhood \cite{ref34} and vehicular distances \cite{ref35}, to represent interactions. This representation cannot accurately reflect the influence among vehicles' motion states, particularly in a 2D highway context where the longitudinal distances between neighboring vehicles are typically more pronounced than the lateral ones. Last but not the least, none of the current methodologies addresses the impact of interactions that could potentially occur in the future. The historical relationship between vehicles does not inherently reflect the interaction in the future \cite{ref58}. As illustrated in Figure \ref{fig_2}, while the target vehicle and vehicle 2 display a weak relationship in historical observations, they will highly interact in the future. This underscores the challenge of representing the future states and interactions based solely on historical trajectories. Predicting the future motion states of the target vehicle and surrounding vehicles, coupled with a thorough analysis and incorporation of the influence of these future states, is vital for accurate prediction of the target vehicle's future trajectory.

To enhance the representation of vehicle interactions and integrate the influence of surrounding vehicles’ motion states on the target vehicle, we propose constructing a graph with a dynamic adjacency matrix. This matrix aims to deeply model interactions between the target vehicle and its neighbors, considering factors such as neighborhood, distance, and associated risks. The longitudinal and lateral potential risk evaluations are derived from the relative distances and velocities between vehicles, assessing the potential collision risks in both directions. Moreover, by leveraging an advanced graph neural network, we effectively capture spatial and temporal dependencies across varied timestamps, embedding historical states and predicting future motion states of the target and surrounding vehicles. This methodology allows for a refined estimation of future interactions and seamlessly integrates these future states into the trajectory predictions of the target vehicle.

\textbf{Multi-modality}. The complex interaction between vehicles contributes to the uncertain and unpredictable nature of real-world traffic scenarios, posing a substantial challenge in accurately forecasting a single future trajectory with high confidence \cite{ref36}. Various factors such as personalized driver characteristics and physical and psychological factors result in diverse behaviors and reactions among drivers for the same driving situation \cite{ref5, ref37}. Hence, modeling the multi-modality of driving behaviors is crucial to generate multiple possible trajectories. Such behavioral multi-modality is an inherent feature of vehicle movement \cite{ref4} and is frequently characterized by the introduction of supplementary latent variables. Existing models addressing this issue can be classified into two categories, distinguished by the presence or absence of explicit semantics or interpretability in the latent variables. 

Specifically, the semantics of driving intentions refer to the purpose behind the motion and behaviors of a driver on the road. It involves understanding what a driver intends to do or achieve while driving, which is crucial for safe and efficient traffic flow. In the first category, the latent variable exhibits clear semantics to represent driver intention. Studies by \cite{ref5, ref19} identifies driving maneuvers as latent variables, covering three lateral and two longitudinal behaviors. \cite{ref17}'s method employs attention heads for trajectory creation in specific scene encodings. These maneuvers are first categorized, then utilized in creating multi-modal trajectory prediction models via Gaussian distribution. Some models assign trajectories based on pre-clustered anchors rather than predetermined maneuvers. Numerous works predict goals using vector anchors and high-definition maps \cite{ref50, ref51}. Additionally, \cite{ref38} incorporates different lane features for predictions. However, most studies directly apply a simple concatenation operation to combine the features of maneuver or lane and encoded context which are insufficient for generating diverse predictions. In the second category, the latent variable is devoid of explicit semantics. This is majorly driven by progress in generative deep learning, particularly Variational Autoencoder (VAE) \cite{ref39} and Generative Adversarial Network (GAN) \cite{ref31, ref40}. GANs incorporate diverse factors in generators \cite{ref53, ref54, ref55} with discriminators evaluating latent details. VAEs employ encoder-decoder structures, refining multi-modal trajectory predictions \cite{ref56, ref57}. Such models deploy latent random variables for crafting varied, multi-modal trajectories, typically adding noise from latent distributions to the encoded features for stochastic outcomes. Major challenges are the lack of interpretability and the difficulty of determining optimal sample sizes and probabilistic assignments for these trajectories.

In this work, we address interpretability by treating multi-modal behaviors as latent variables with explicit semantics, effectively representing potential driving intentions. The backbone of the feature fusion technique is derived from the study by \cite{ref23}. Instead of producing a solitary multi-modal prediction for the entire future temporal horizon, we formulate a probabilistic framework which can predict the possible semantic intention at each distinct future timestamp. By directly bridging the intrinsic intention with feasible trajectories via intention-specific feature combinations, this approach successfully accomplishes multi-modal prediction.

Based on the discussion of the two major challenges, this work proposes a Graph-based Interaction-aware Multi-modal Trajectory Prediction (GIMTP) framework which aims to study the deep interactions between the vehicles and provide multiple predictions in probabilistic manner. We construct a dynamic graph embedding for each vehicle group on the highway to capture the intuitive and deep interaction, with a particular emphasis on potential risk factors. The graph neural network-based interaction encoder maps the historical motion states of the vehicle group to both a historical embedding and an inferred future-guided future embedding. These embeddings facilitate intention predictions at each time step in the future horizon. Distinct intentions are incorporated through a feature fusion process within the multi-modal decoder, enabling the generation of multiple possible future trajectories based on the intention. The effectiveness of the proposed model is evaluated with multiple experiments and case studies. Three main contributions are summarized as follows:
\begin{enumerate}

\item Represent the motion states of vehicles through a dynamic graph, thereby encapsulating the interactions between vehicles with comprehensive consideration of potential risk factors.
\item Propose a future-guided graph embedding that maps historical motion states onto future states, elucidating the correlation between the future motion states of neighboring vehicles and the future motion states of the target vehicle.
\item Provide predictions for both longitudinal and lateral driving behaviors in a multi-modal fashion, with related potential future trajectories designated respective probabilities.
\end{enumerate}

The remainder of this paper is structured as follows: Section II outlines the methodology employed in the study. Section III details the experimental setup, while Section IV provides an in-depth analysis of the obtained results. Section V concludes this paper.

\section{Methodology}
\subsection{Problem Statement}
This study endeavors to forecast the trajectory of the target vehicle in a multi-lane highway scenario by leveraging the historical motion states of the target vehicle and its surrounding vehicles. The target vehicle is assumed to be a CAV and its surrounding vehicle can be the combination of human-driven vehicles and CAVs. Within the scope of our research, we aim to analyze the interaction between the target vehicle and its surrounding vehicles, as well as the multi-modality of the target vehicle’s driving behavior. 

Mathematically, the task of vehicle trajectory prediction can be formulated as the prediction of the probability distribution of the target vehicle’s future trajectory position based on the observed historical motion information of the target vehicle and its surrounding vehicles. The historical states of the vehicle group in a historical time horizon $[1,\dots,T]$ can be denoted as $X_{1:T}=\{X_1,X_2,\dots,X_T\}$. Each historical state $X_t$ at time step $t\in\{1,\dots,T\}$ represents the union set of the historical states of target vehicle and its surrounding vehicles. Thus, $X_t=\{\mathbf{x}_t^0,\mathbf{x}_t^1,\dots,\mathbf{x}_t^N\}$, where $\mathbf{x}_t^i$ represents the historical state of vehicle $i$, for all $i\in\{0,1,\dots,N\}$ and all $t\in\{1,\dots,T\}$. In this paper, superscript refers to vehicle indices, where $i=0$ especially for the target vehicle, and the subscript refers to time steps. The state $\mathbf{x}_t^i$ associated with vehicles $i$ could include its longitudinal and lateral positions and velocity. 

Suppose the current time step is at step $T$, then the predicted states of the vehicle group in a future time horizon $[T+1,\dots,T+F]$ can be denoted as $\hat{Y}_{T+1:T+F}=\{\hat{Y}_{T+1},\hat{Y}_{T+2},\dots,\hat{Y}_{T+F}\}$. Each predicted future state $\hat{Y}_{T+f}$ at time step $T+f$ only composed of the predicted state of the target vehicle. Thus $\hat{Y}_{T+f}=\{\hat{\mathbf{y}}_{T+f}^0\}$, where $\hat{\mathbf{y}}_{T+f}^0$ denote the future state of the target vehicle at time step $T+f$ for all $f\in\{1,\dots,F\}$. The state $\hat{\mathbf{y}}_{T+f}^0$ consists of the longitudinal and lateral positions of the target vehicle at future time step $T+f$. 

Notice that the coordinates of all vehicles in the vehicle group are expressed in a reference frame where the origin is the position of the target vehicle at timestamp $T$. The input to the model consists of $X_{1:T}=\{X_1,X_2,\dots,X_T\}$ during the past $T$ time steps and the output of the model is a probability distribution $P (\hat{Y}_{T+1:T+F} |X_{1:T})$ over the next $F$ time steps. In this work, the distribution of $\mathbf{y}_{T+f}^0$ is parameterized as a bivariate Gaussian distribution with mean $(\mu_{T+f,x},\mu_{T+f,y})$, variance $(\sigma_{T+f,x}^2, \sigma_{T+f,y}^2)$, and correlation coefficient $\rho_{T+f}$, where the subscript $x$ stands for the longitudinal position and $y$ stands for the lateral position.

Thus, the trajectory prediction problem can be summarized as follows. Given the states $X_{1:T}$ of all the vehicles in a vehicle group over a past time horizon $[1,\dots,T]$, the objective is to train a model $\mathbf{H}(\cdot)$ to predict the trajectory distributions $\hat{Y}_{T+1:T+F}$ of the target vehicle that approximates the ground truth trajectory $Y$ in the future time horizon $[T+1,\dots,T+F]$:
\begin{equation}
    \hat{Y}_{T+1:T+F} = \mathbf{H} (X_{1:T})
\end{equation}

\subsection{Model Architecture}
To achieve precise trajectory prediction in dense traffic, it is essential to capture the complex temporal and social correlations between the target vehicle and other vehicles. Therefore, we propose a model comprising the following modules:
\begin{itemize}
    \item Dynamic Graph Embedding Module, which transforms the original vehicle motion states to a dynamic graph embedding considering the neighborhood, the distance, and the potential risk between the vehicles.
    \item Interaction Encoder, which captures the interaction between the vehicles by adopting a Diffusion Graph Convolution Network (DGCN) architecture.
    \item Intention Predictor, which maps the historical encoding and the guiding future trajectories to both the lateral and longitudinal intentions in the future time horizon.
    \item Multi-modal Decoder, which fuses the predicted intentions with the latent space and provides multiple future trajectory distributions with possibilities. 
\end{itemize}

\begin{figure}[!h]
\centering
\includegraphics[scale=0.34]{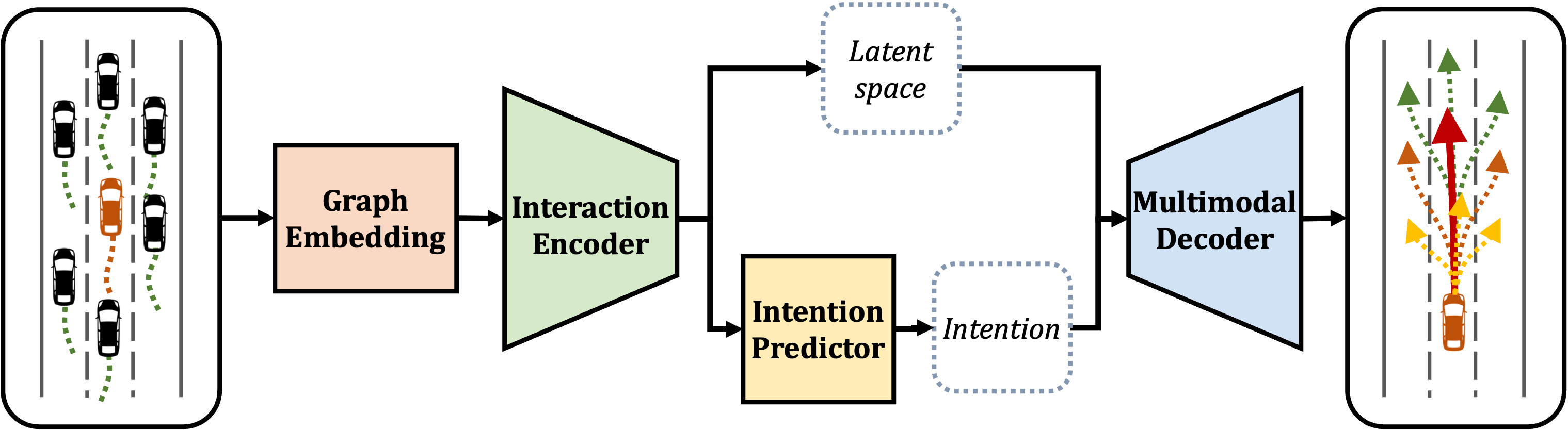}
\caption{Framework architecture.}
\label{fig_3}
\end{figure}

\subsubsection{Dynamic Graph Embedding Module}
\begin{figure}[!h]
\centering
\includegraphics[scale=0.26]{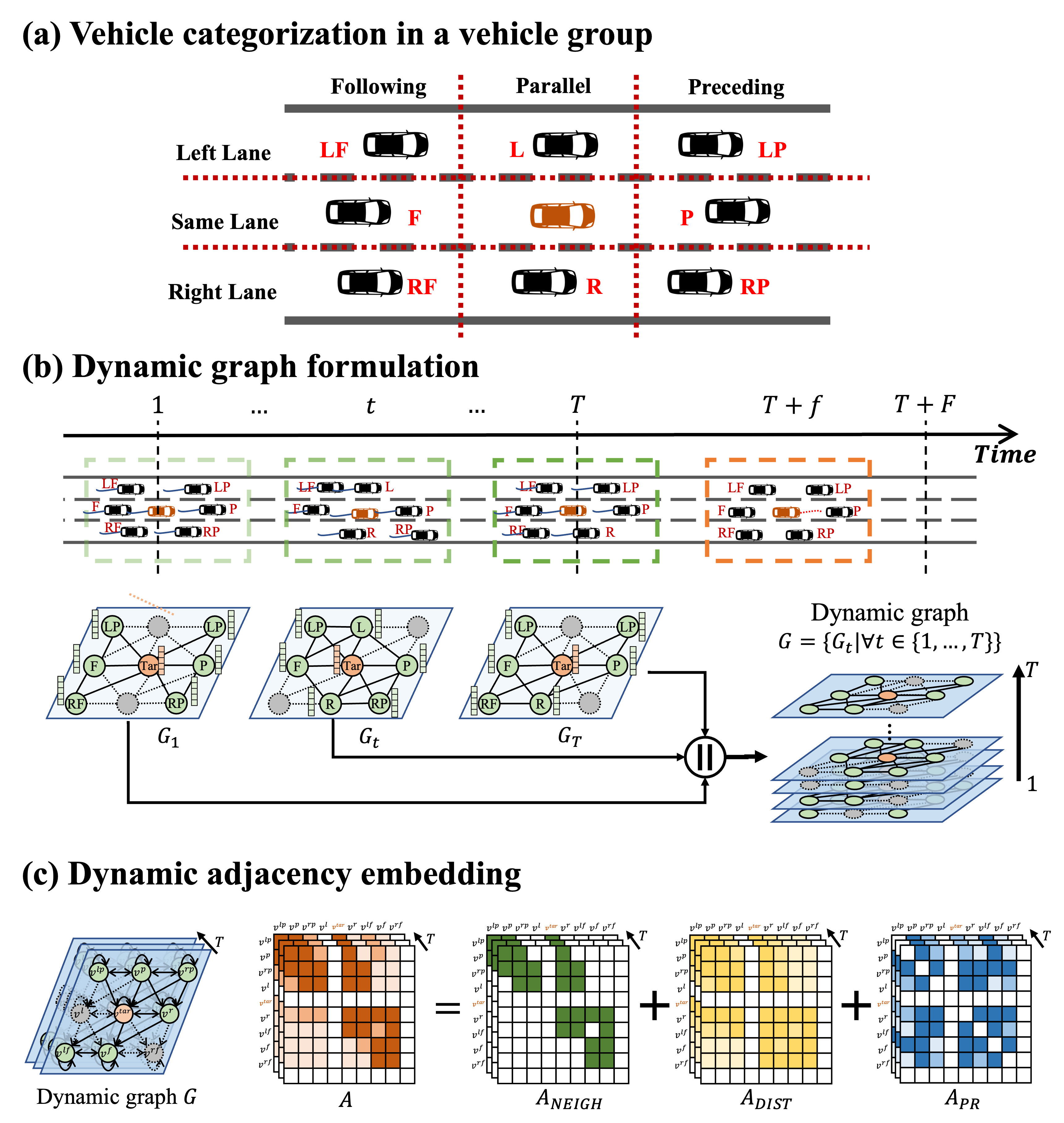}
\caption{Dynamic graph embedding of vehicle group’s motion states.}
\label{fig_4}
\end{figure}

To provide accurate trajectory prediction of the target vehicle, the correlation and interaction between the target vehicle and its surrounding vehicles is critical. In the vehicular group depicted in Figure \ref{fig_4}(a), the surrounding vehicles can be categorized longitudinally into the preceding vehicles, the parallel vehicles, and the following vehicles, and laterally into the left vehicles, the same-lane vehicles, and the right vehicles. A dynamic spatial-temporal graph $G_{1:T}=\{G_t |\forall t \in\{1,\dots,T\}\}$ is constructed to represent the movement state of the vehicle group from step 1 to step $T$. Specifically, the graph $G_t=(V_t,E_t;X_t,A_t)$ represents the formulation of the vehicle group at time step $t$. Each node $v_t^i$ in the node set $V_t=\{v_t^i |\forall i\in\{1,\dots,N\}\}$ represents each vehicle $v^i$ in the vehicle group at step $t$. The edge set $E_t$ generally represents the influence between the vehicles. A zero value for edge $e^{ij}$ indicates there is no interaction between node $v^i$ and node $v^j$ for $i,j\in\{1,\dots,N\}$. $X_t\in \mathcal{R}^{N \times C}$ is the feature matrix of the vehicle group, where $C$ is the number of features. 

Vehicle motion can exert diverse effects on nearby vehicles. For instance, a sudden lane change or a sudden change in speed might induce adjacent vehicles to reduce speed or alter their path, whereas it may exert negligible influence on more distant vehicles. To quantitatively capture these interactions, we define a weighted adjacency matrix $A_t = \left(A_t^{ij}\right) \in \mathcal{R}^{N \times N}$ as a weighted adjacency matrix for the graph $G_t$, where each element $A_t^{ij}$ denotes the intensity of interactions between vehicles $v^i$ and $v^j$. This dynamic adjacency matrix is constructed with considerations for neighborhood, distance, and potential risks to accurately describe vehicle interactions. Subsequent sections delve into these factors in detail.

\textbf{Adjacency dependent on neighborhood}: From an intuitive perspective, a vehicle's motion states can have a significant impact on the movement of its surrounding vehicles, particularly those that are in close proximity. To capture this relationship, an adjacency matrix that considers the immediate neighborhood is defined as follows:
\begin{equation}
    A_{t,NEIGH}^{ij} = 
    \begin{cases}
        1, & \text{if } v_t^i \text{ and } v_t^j \text{ are adjacent}\\
        0, & \text{o.w.}
    \end{cases}
\end{equation}

\textbf{Adjacency dependent on distances}: Intuitively, two vehicles with closer distances may have stronger effects on each other, we use a function decaying over distance to measure the weight between two vehicles so that closer vehicles have higher weights:
\begin{equation}
    A_{t,DIST}^{ij} = \exp\left(-\left(\frac{dist(v_t^i,v_t^j )}{\sigma_{dist}}\right)^2\right)
\end{equation}
where $dist(v_t^i,v_t^j)$ is the Euclidean distance between the node $v^i$ and $v^j$ at step $t$, and $\sigma_{dist}$ is the standard deviation of all the distances between every pair of nodes. 

\textbf{Adjacency dependent on potential risks}: Potential field, as a vital component for capturing social interactions \cite{ref41, ref42}, which has been extensively applied in autonomous vehicle path planning \cite{ref43}, has proven its potential in trajectory prediction \cite{ref44}. We are considering utilizing the kinetic energy of the vehicles to represent the risk to better address the potential collision between the vehicles. When a vehicle collides with another, its kinetic energy is released, transmitted, or converted, which is an anomalous energy transfer process according to the theory of energy transfer \cite{ref45}. Thus, the potential traffic risk can be described as follows:

\begin{equation}
    E^i = \frac{1}{2} m^i (s^i)^2 = \frac{1}{2} m^i s^i \cdot \frac{s^i-0}{\Delta d^i}\cdot \Delta d^i
\end{equation}
where $E^i,m^i,s^i$ are the kinetic energy, mass, and velocity of vehicle $v^i$, respectively. $\Delta d^i$ denotes the distance between vehicle and another position in the traffic environment. Since $E^i=F^i \Delta d^i$, the equivalent force $F^i$ caused by vehicle $v^i$ can be represented as
\begin{equation}
    F^i = \frac{1}{2} m^i s^i \cdot \frac{s^i-0}{\Delta d^i}
\end{equation}

In a car-following scenario including a follower vehicle $v^i$ and a leader vehicle $v^j$, the traffic risk and the corresponding internal equivalent force between the two vehicles can be represented as
\begin{equation}
    E^{ij} = \frac{1}{2} m^i s^i \frac{s^i-s^j}{|d^i-d^j|} |d^i-d^j|
\end{equation}
\begin{equation}
    F^{ij} = \frac{1}{2} m^i s^i \frac{s^i-s^j}{|d^i-d^j|}
\end{equation}
where $s^j,d^j$ are the velocity and longitudinal position of vehicle $v^j$. The term $(s^i-s^j)/|d^i-d^j|$ indicates the relative velocity between vehicle $v^i$ and vehicle $v^j$ divided by their relative distance. Note that the risk exists if the relative velocity is positive. This strategy can be extended to consider the risk in both the longitudinal dimension and lateral dimension for the vehicle group. At each timestep $t$, the longitudinal equivalent force and lateral equivalent force between two vehicles $v^i$ and $v^j$ can be represented as
\begin{equation}
    F_{t,y}^{ij} = 
    \begin{cases}
        0, & \text{if } s_{t,y}^i - s_{t,y}^j \le 0 \\
        \frac{1}{2} m^i s_{t,y}^i \frac{s_{t,y}^i-s_{t,y}^j}{|d_{t,y}^i-d_{t,y}^j|}, & \text{o.w.} \\
    \end{cases}
\end{equation}
\begin{equation}
    F_{t,x}^{ij} = 
    \begin{cases}
        0, & \text{if } s_{t,x}^i - s_{t,x}^j \le 0 \\
        \frac{1}{2} m^i s_{t,x}^i \frac{s_{t,x}^i-s_{t,x}^j}{|d_{t,x}^i-d_{t,x}^j|}, & \text{o.w.} \\
    \end{cases}
\end{equation}
where $s_{t,x}^i,d_{t,x}^i$ denote the lateral speed and position of vehicle $v^i$ at time step t, and $s_{t,y}^i,d_{t,y}^i$ denote its longitudinal velocity and position. The superscript represents the information of vehicle $v^j$ at the same timestep. The resultant force $F^{ij}$ can be represented as 
\begin{equation}
    F_t^{ij} = \sqrt{(F_{t,x}^{ij})^2+(F_{t,y}^{ij})^2}
\end{equation}

Thus, the adjacency matrix incorporating the potential risk can be represented as
\begin{equation}
    A_{t,PR}^{ij} = \tanh\left(\frac{F_t^{ij}}{\sigma_F}\right)
\end{equation}
where $\tanh(\cdot)$ is a hyperbolic tangent function, and $\sigma_{F}$ is the standard deviation of all the forces between every pair of nodes. 

\textbf{Dynamic adjacency embedding}: The dynamic adjacency matrix $A \in \mathcal{R}^{T\times N \times N}$ is the concatenation of $A_1,\dots,A_T$. For $\forall t\in[1,\dots,T]$, $A_t$ is the normalized summation of the three adjacency matrices we defined considering neighborhood, distance, and potential risks factors:
\begin{equation}
    A_t = normalize(A_{t,NEIGH}+A_{t,DIST}+A_{t,PR})
\end{equation}
such dynamic adjacency matrix $A$ reflects the intuitive and comprehensive way to capture the dynamic interactions between vehicles. $A$ is further normalized in range between 0 and 1.

It’s worth noting that the graph $G$ is sparse and time-varying over the time horizon $[1,T]$. Since no single vehicle is assigned to any relative position in the vehicle group over a time horizon, there does not always exist vehicles in each of the eight positions, resulting in the sparsity of the graph. Furthermore, since the vehicles are moving and the distance between the vehicles changes over time, the adjacency matrix keeps varying. A surrounding vehicle may change its relative position to the target vehicle. For example, a vehicle on the left of the target vehicle may accelerate and change lanes so that it moves from node $v^l$ to node $v^p$. A vehicle is also free to leave the vehicle group, and new vehicles may join the group. Such sparsity and variation in time characterize the microscopic interaction between vehicles.

\subsubsection{Graph Interaction Encoder}
Inspired from \cite{ref52}, we adopt the a Diffusion Graph Convolutional Networks (DGCN) module to characterize the bidirectional dependencies between nodes in the graph embedding. Denoting the DGCN layer as $DGCN^L(\cdot)$, the diffusion convolution is operated over the graph signal including the diffusion process and its reverse process:
\begin{equation}
    \begin{split}
        H_{l+1}
    &= DGCN^L(H_l) \\
    &= \sum_{k=1}^{K} \left(T_k(\bar A_f) \cdot H_l \cdot \Theta_{f,l}^k + T_k(\bar A_b) \cdot H_l \cdot \Theta_{b,l}^k \right)
    \end{split}
\end{equation}
where $H_{l+1}$ is the output of the $l$-th layer; the masked feature matrix $X$ is the input of the first layer. The computation block mapping $H_l$ to $H_{l+1}$ is denoted as $DGCN(\cdot)$. The weighted adjacency matrix $W$ records the relation between the nodes and the message passing directions. The forward transition matrix $\bar A_f=A/rowsum(A)$ can capture the dependency of the downstream nodes, and the backward transition matrix $\bar A_b=A^T/rowsum(A^T)$ can capture the dependency of the upstream nodes. $T_k(\cdot)$ is a Chebyshev polynomial and k is the order, which could approximate the convolution process of the $k$-th layer neighbors for each node; $T_k(X)=2X \cdot T_{k-1}(X) - T_{k-2}(X)$. $\Theta_{b,l}^k$ and $\Theta_{f,l}^k$ are the learning parameters of the $l$-th layer which give weights to the input information. For the target vehicle in the vehicle group, the forward diffusion process could capture the influence of the downstream vehicles and its reverse process could capture the influence of the upstream vehicles.

The calculation process of the three-layer DGCN module in the encoder can be summarized as follows:
\begin{equation}
    H_1 = DGCN^L_1(X)
\end{equation}
\begin{equation}
    H_2 = \sigma(DGCN^L_2(H_1)) + H_1
\end{equation}
\begin{equation}
    H_o = DGCN^L_3(H_2)
\end{equation}
where $H_1,H_2$ are the output of the first and second DGCN layer, respectively. The hidden state $H_o$ is encoded feature matrix of the encoded graph $\tilde G$, which is the output of the DGCN module.

\begin{figure}[!h]
\centering
\includegraphics[scale=0.28]{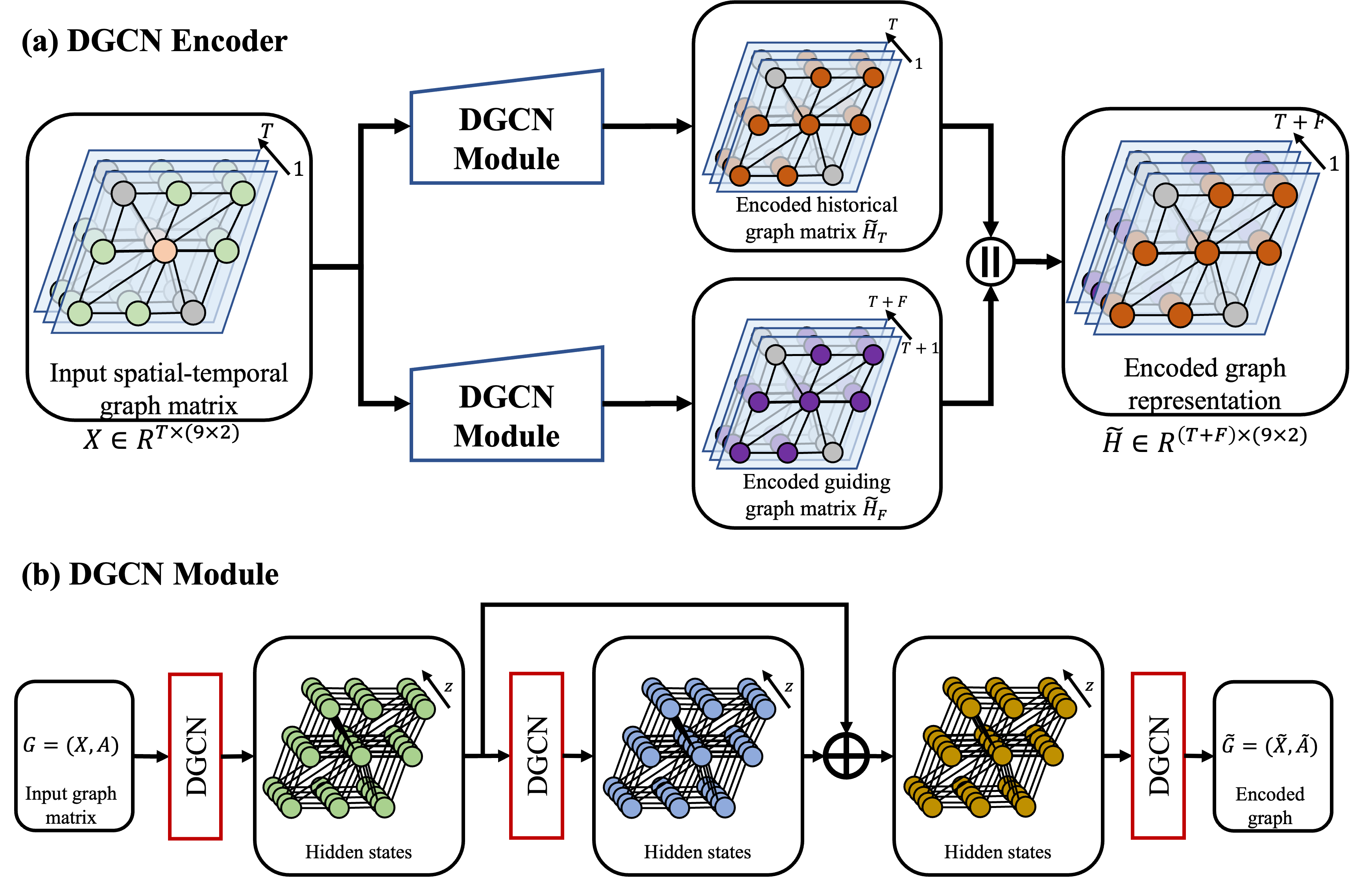}
\caption{Intention Encoder.}
\label{fig_5}
\end{figure}

We adopt two DGCN modules which contains a three-layer DGCN architecture to capture the bidirectional propagation among the vehicle group. One module encodes the historical states of the vehicle group in historical time horizon $[1,\dots,T]$ to a historical graph embedding as $\tilde H_T$, while the other maps the vehicle group’s historical states to the states in the future time horizon $[T+1,\dots,T+F]$ as a future-guided graph embedding denoted as$\tilde H_F$. 
\begin{equation}
    \tilde H_{T} = DGCNEnc_H(X)
\end{equation}
\begin{equation}
    \tilde H_{F} = DGCNEnc_F(X)
\end{equation}
\begin{equation}
    \tilde H = [\tilde H_T, \tilde H_F]
\end{equation}
The output graphs of the two modules are concatenated to provide an integrated embedding $\tilde H$ of the motion states in the time window containing both historical steps and future steps.

\subsubsection{Intention Predictor}
The actual trajectory of a vehicle in real traffic scenes is often uncertain due to the complexity and diversity of possible driving intentions. Noticing that for multi-lane freeway, the type of driver’s intention is generally limited, we distinguish lanes keeping (LK), left lane change (LLC) and right lane change (RLC) as three lateral intention classes, as well as constant speed (CS), acceleration (ACC), and deceleration (DEC) as three longitudinal intention classes. Denote the intention set as $M=[m^{lat}, m^{lon}]$, where $m^{lat} = [m^{LK},m^{LLC},m^{RLC}]\in \mathcal{R}^{ 3 \times F}$ and $m^{lon} = [m^{CS},m^{ACC},m^{DEC}]\in \mathcal{R}^{ 3 \times F}$. One-hot encoding is used for both $m^{lat}$ and $m^{lon}$. To this end, we introduce intention recognition to obtain the probability of different intentions which in turn assists the implementation of multi-modal trajectory prediction described in the next module. We use the same intention labeling method following the method in \cite{ref8}. A fully connected layer with softmax activation function is utilized to calculate the probability of lateral intention and longitudinal intention classes as follows:
\begin{equation}
    h_{m} = MLP_O (MLP_V (\tilde H; W_{m}))
\end{equation}
\begin{equation}
    P(m^{lat}) = softmax(LatMLP(h_{m}))
\end{equation}
\begin{equation}
    P(m^{lon}) = softmax(LonMLP(h_{m}))
\end{equation}
where the fully connected layer $MLP_V$ aggregates the hidden states of the vehicle groups and the layer $MLP_O$ maps the input time horizon to the output time horizon.

\begin{figure}[!h]
\centering
\includegraphics[scale=0.32]{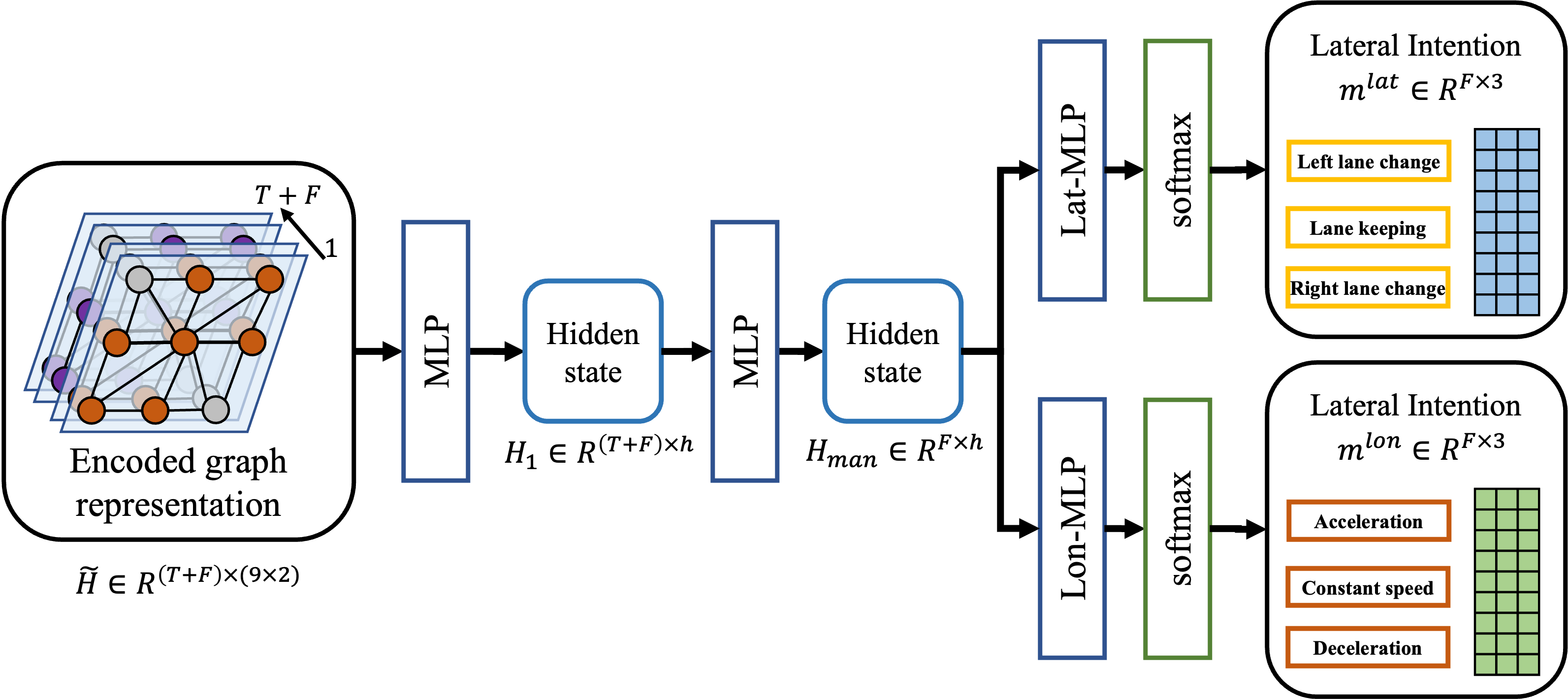}
\caption{Intention Predictor.}
\label{fig_6}
\end{figure}
Intuitively, the feature vectors acquired from the encoder at different historical timestamps contribute unequally with respect to different prediction horizons.

\subsubsection{Multi-modal Decoder}
Intuitively, the feature vectors obtained from the encoder at various historical time points contribute unequally with regard to distinct prediction horizons. For instance, the position of a vehicle at time stamp $T + 1$ is more closely related to its location at time stamp $T$ than to that at time stamp $1$ due to the inherent continuity of motion. The predicted trajectory under differing intentions should vary, thereby necessitating unique features as input. However, it remains uncertain which feature vectors supplied by the encoder hold greater relevance than others for accurate prediction concerning a specific horizon. To tackle this challenge, we propose an intention-specific feature fusion approach that explicitly contemplates the significance of encoded feature representations by adaptively combining features at disparate historical time stamps.

Given the encoded feature vectors  $\tilde H = [\tilde h_1, \tilde h_2, \dots, \tilde h_T]$, acquired from the encoder, a sequence of weight vector $u_{t'}^m=[u_{1,t'},u_{2,t'},\dots,u_{T,t'}]$ for each intention $m$ from the six intentions $[m^{lat}, m^{lon}]$ at $t'=[T+1,\dots,T+F]$ is respectively used to combine $\tilde h_t$ at different timestamps as follows:
\begin{equation}
    v_{t'}^m=\sum_{t=1}^{T+F} u_{t,t'}^m  \tilde h_t
\end{equation}

The above weight vectors for each intention of the lateral intentions and longitudinal intentions can be organized in matrix form as
\begin{equation}
    W_m = 
    \begin{bmatrix}
        u_{1,1}^m & \dots & u_{1,F}^m \\
        \vdots & \dots & \vdots  \\
        u_{T,1}^m & \dots & u_{T,F}^m \\
        u_{T+1,1}^m & \dots & u_{T+1,F}^m \\
        \vdots & \dots & \vdots  \\
        u_{T+F,1}^m & \dots & u_{T+F,F}^m \\
    \end{bmatrix}
    \in \mathcal{R}^{(T+F)\times F}
\end{equation}

Obviously, the value $u_{t,t'}^m$ of implies the contribution of feature vector $\tilde h_t$  when predicting intentions for $t'=[T+1,…,T+F]$. More important features tend to have larger weights and vice versa.

\begin{figure}[!h]
\centering
\includegraphics[scale=0.32]{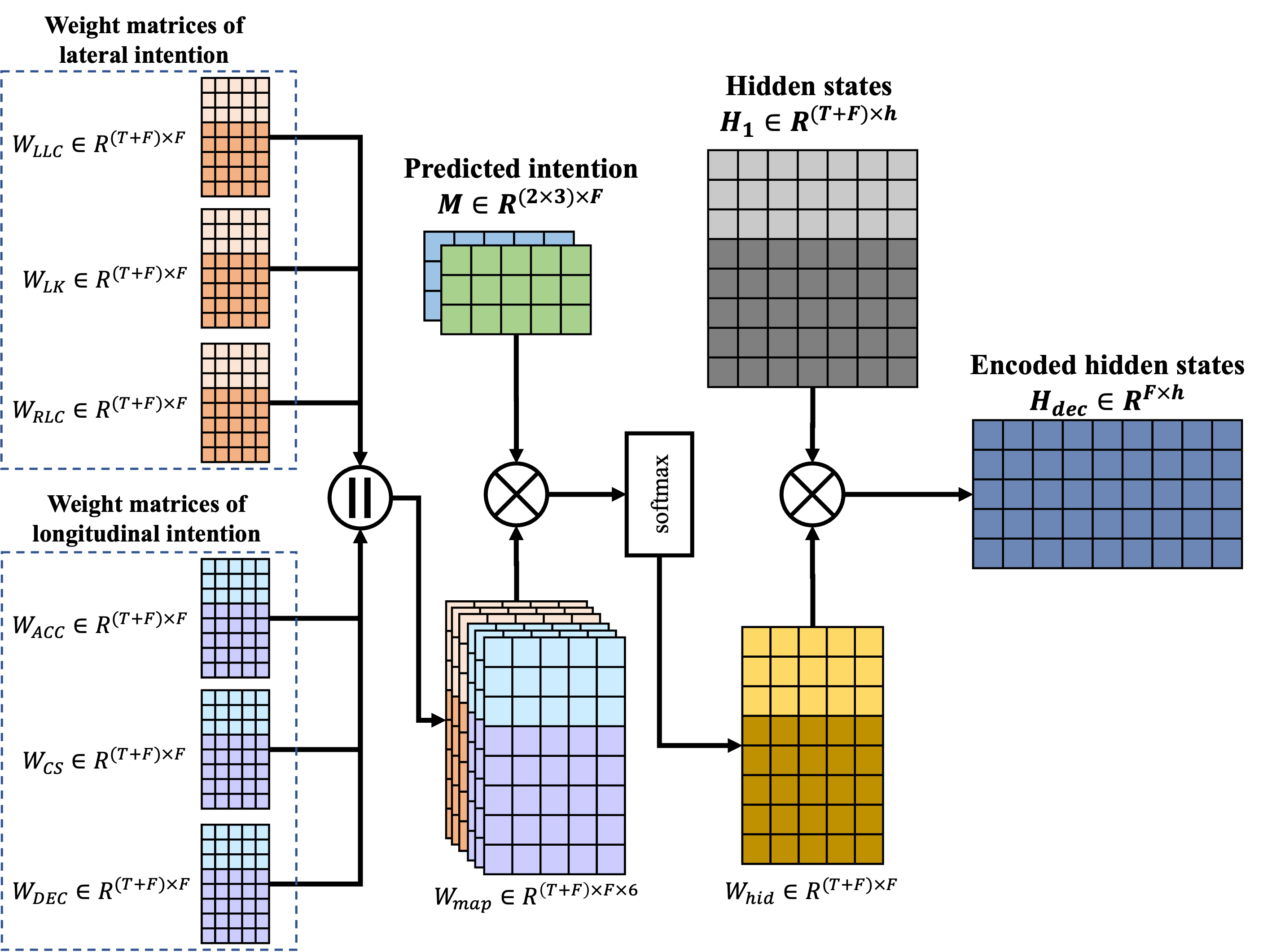}
\caption{Intention feature fusion.}
\label{fig_7}
\end{figure}

The weight matrices of the six intentions are stacked as $W_{map} \in \mathcal{R} ^{(T+F) \times F \times 6}$. With the predicted intentions matrix $M \in \mathcal{R}^{6\times F}$ and the latent space $\tilde H$, the resulting encoded hidden states $H_{dec}$ for the Multi-modal Decoder can be calculated as:
\begin{equation}
    W_{hid} = softmax(W_{map} \cdot M)
\end{equation}
\begin{equation}
    H_{dec} = W_{hid} \cdot \tilde H
\end{equation}
where $W_{hid}$ is the interim hidden states. To estimate the probability distribution of trajectory prediction, we obey the total probability theorem and factorize $P(Y|X)$ as follows:
\begin{equation}
    P(Y|X) = \sum_m P_\theta (Y|X,m) P(m|X)
\end{equation}
where $\theta=[\theta_{T+1},\theta_{T+2},…,\theta_{T+F}]$ denotes the parameters of bivariate Gaussian distribution of the target vehicle position at each timestamp in the future. Each $\theta_{t'}=\{\mu_{t',x}, \mu_{t',y}, \sigma_{t',x}, \sigma_{t',y}, \rho_{t'}\}$ consists of the mean, variance, and correlation coefficient of the vehicle lateral position and longitudinal position thus reflecting the uncertainty of the prediction. 

\begin{figure}[!h]
\centering
\includegraphics[scale=0.38]{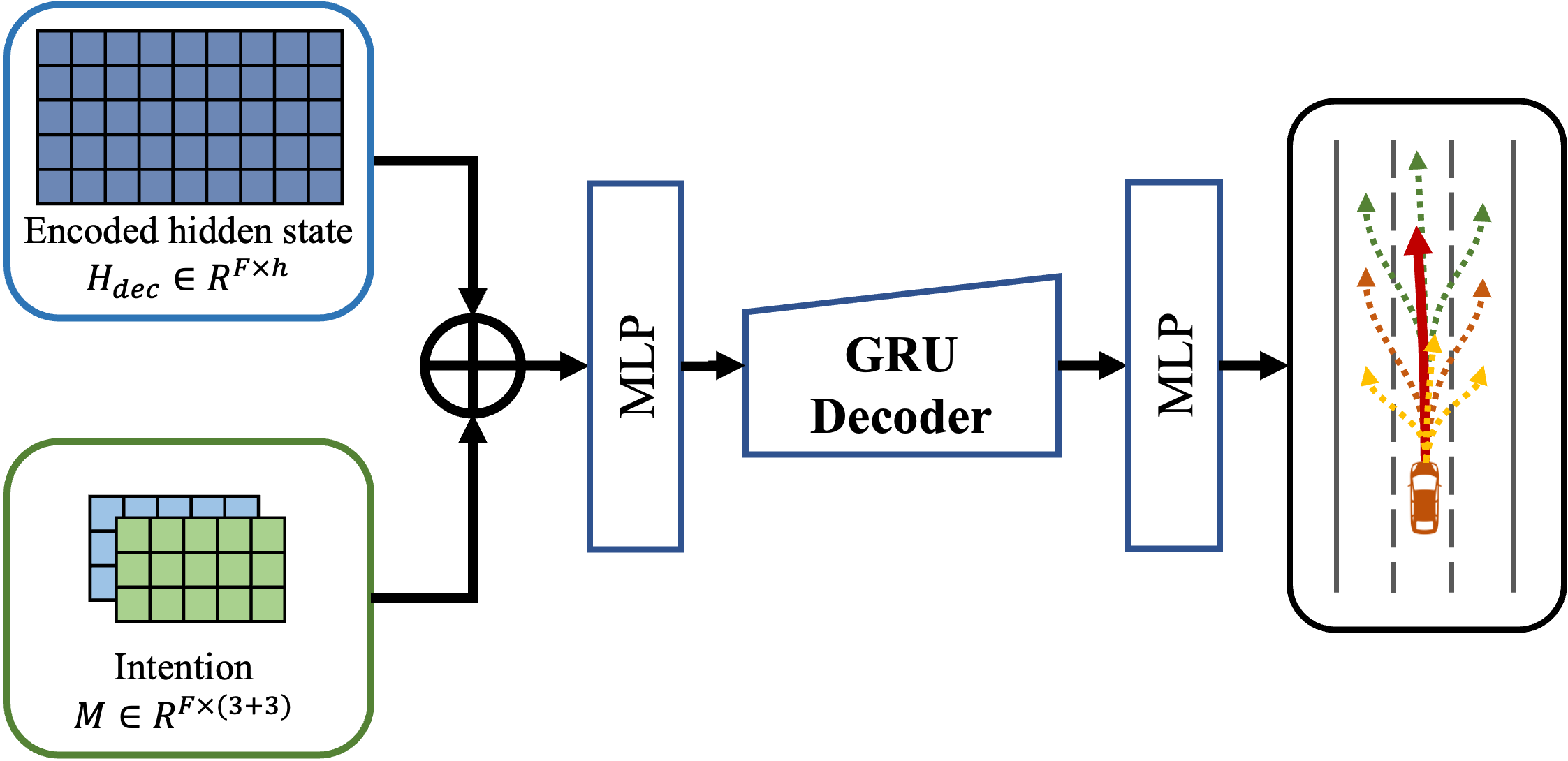}
\caption{Multimodal Decoder.}
\label{fig_8}
\end{figure}

In order to predict the probability distribution of the future trajectory based on different intentions, the intention-specific feature $v_{t'}^m$ in $H_{dec}$ and the prediction probability of intention classes are concatenated and then fed into a fully connected layer with weight $W_{fc}$ as follows:
\begin{equation}
    e_{t'}^m = MLP_{D1}(v_{t'}^m \oplus P(m); W_{fc})
\end{equation}

Additionally, considering the temporal continuity of predicted trajectory, the encoded representation $e_{t'}^m$ at each timestamp is sequentially fed into a decoder GRU which outputs the required predictive distribution parameters $\theta_{t'}$ for vehicle motion, that is
\begin{equation}
    h_{t'} = GRU(e_{t'}^m, h_{t'-1}; W_{GRU})
\end{equation}
\begin{equation}
    \theta_{t'} = MLP_{D2}(h_{t'}, W_{de})
\end{equation}
where $W_{GRU}$ denotes the weight in GRU cell, $W_{de}$ is the weight used to convert hidden state $h_{t'}$ into the five parameters of bivariate Gaussian distribution.

\subsection{Training Metrics}
We adopt a two-stage training methodology, leveraging varied loss function components to enhance the learning of network parameters. Specifically, during the initial five epochs, we aim to minimize the Mean Square Error (MSE) loss, given its innate ease of optimization
\begin{equation}
    MSE(\hat Y;Y) =\frac{1}{F} \sum_{t'=T+1}^{T+F} \left( \left(\mu_{t',x}-x_{t'}\right)^2 + \left(\mu_{t',y}-y_{t'}\right)^2 \right) 
\end{equation}
where $Y =(x_{t'},y_{t'})$ is the ground truth position of the target vehicle at timestamp $t$, $\hat Y= (\mu_{t',x}, \mu_{t',y})$ is the predicted position by the model. Then, beginning from the sixth epoch, we switch to the following negative log likelihood (NLL) loss to further train the network
\begin{equation}
    NLL = -\log \left( \sum_m P_\theta (Y|X,m) P(m|X) \right)
\end{equation}

For each training sample, only one intention class is assigned and therefore we choose to optimize the following NLL loss function instead
\begin{equation}
\begin{split}
    NLL &= -\log (P_\theta (Y|X,m) P(m|X)) \\
    &= -\log P_\theta (Y|X,m) - \log P(m|X) \\
    &= NLL_{traj} + NLL_{m}
\end{split}
\end{equation}
To minimize the loss of intention recognition related terms $NLL_m$, we adopt the following cross entropy loss
\begin{equation}
    NLL_{m} = - \sum_{m} y^m \log P(m|X)
\end{equation}
where $y^m$ denotes the ground truth intention. For the trajectory related term $NLL_{traj}$, according to bivariate Gaussian distribution, we have
\begin{equation}\label{eq35}
\begin{split}
    &NLL_{traj}(\hat Y;Y) \\
    &= \sum_{t'=T+1}^{T+F} \left( \log \left( 2\pi\sigma_{t',x} \sigma_{t',y} \sqrt{1-\rho_{t'}^2} \right. \right.\\
    &\quad  + \frac{1}{2(1-\rho_{t'}^2)} \left( \frac{(\mu_{t',x}-x_{t'})^2}{\sigma_{t',x}^2} \right.\\
    &\quad \left.\left.\left. - \frac{(\mu_{t',x}-x_{t'})(\mu_{t',y}-y_{t'})}{\sigma_{t',x}^2} + \frac{(\mu_{t',y}-y_{t'})^2}{\sigma_{t',y}^2} \right) \right) \right)
\end{split}
\end{equation}

Thus, the overall two-stage loss function $\mathcal{L}$ is defined as:
\begin{equation}
    \mathcal{L} = \\
    \begin{cases}
        MSE(\hat Y; Y) + \alpha \cdot NLL_m \\
        \quad \quad + \beta \cdot MSE(\tilde H_F; H_F), &\text{if epoch} \le 5\\
        NLL_{traj}(\hat Y; Y) + \alpha \cdot NLL_m \\
        \quad \quad + \beta \cdot MSE(\tilde H_F; H_F), &\text{o.w.}\\
    \end{cases}
\end{equation}

The present study employs the PyTorch deep learning framework to instantiate the proposed network architecture. Specifically, the Adam optimizer is leveraged as the optimization algorithm, utilizing an initial learning rate of 0.01 and a decaying factor to train the network in an end-to-end fashion.

\begin{table}[!h]
\caption{Parameter Settings\label{tab:table1}}
\centering
\begin{tabular}{>{\centering\arraybackslash}m{3cm} >{\centering\arraybackslash}m{0.6cm} |>{\centering\arraybackslash}m{3cm} >{\centering\arraybackslash}m{0.6cm}}
\hline\hline
Parameter & Value & Parameter & Value\\
\hline
neuron \# of DGCN & 256 & neuron \# of MLP$_{D1}$ & 128\\
neuron \# of MLP$_V$ & 256 & neuron \# of MLP$_{D2}$ & 128\\
neuron \# of MLP$_O$ & 256 & learning rate & 0.01\\
neuron \# of LatMLP & 256 & decaying factor & 0.95\\
neuron \# of LonMLP & 256 & $\alpha$ & 0.2\\
neuron \# of GRU & 128  & $\beta$ & 0.1\\         
\hline\hline
\end{tabular}
\end{table}

\section{Experiment Settings and Metrics}
\subsection{Data Preparations}
Our study utilizes two datasets. The first is the Next Generation Simulation (NGSIM) \cite{ref46}, \cite{ref47}, offering detailed vehicle trajectory data from eastbound I-80 in the San Francisco Bay area and southbound US 101 in Los Angeles. Collated by the U.S. Department of Transportation in 2015, it captures real-world highway scenarios via overhead cameras at 10Hz. The second is the HighD \cite{ref48}, sourced from drone video recordings at 25 Hz between 2017 and 2018 around Cologne, Germany. Covering approximately 420 m of two-way roads, it documents 110,000 vehicles, both cars and trucks, traveling a total distance 45,000 km.

\subsection{Baseline Models}
We compare our proposed model with the following listed baseline models:
\begin{itemize}
    \item Social-LSTM (S-LSTM) \cite{ref7}: a model in which a shared LSTM is used to encode the raw trajectory data for each vehicle and then the extracted features of different vehicles are aggregated by the social pooling layer.
    \item Convolutional Social-LSTM (CS-LSTM) \cite{ref19}: different from S-LSTM, this model captures the social interaction by stacking convolutional and pooling layers and takes into account the multi-modality based on the predicted intention.
    \item Planning-informed prediction (PiP) \cite{ref49}: this model couples trajectory prediction as well as the planning of the target vehicle by conditioning on multiple candidate trajectories of the target vehicle.
    \item Graph-based Interaction-aware Trajectory Prediction (GRIP) \cite{ref26}: this model uses a graph-based representation for interactions between objects, employs graph convolutional layers for feature extraction, and implements an encoder-decoder LSTM for predictive analysis.
\end{itemize}

\subsection{Evaluation Metrics}
The prediction performance of various methodologies is evaluated using the Root Mean Square Error (RMSE). This metric quantitatively measures the difference between the predicted position, represented as $(\mu_{t',x}^l,\mu_{t',y}^l)$, and the ground truth position, denoted by $(x_{t'}^l,y_{t'}^l)$, at distinct timestamps $t’$ in the prediction horizon $[T+1,T+F]$. 
\begin{equation}
    RMSE= \sqrt{\frac{1}{LF}\sum_{l=1}^L \sum_{t'=T+1}^{T+F} \left((\mu_{t',x}^l-x_{t'}^l)^2 + (\mu_{t',y}^l-y_{t'}^l)^2 \right)}
\end{equation}
The superscript $l$ denotes the $l$-th test sample out of the total number of test samples $L$. In scenarios of multi-modal predictions, our model can produce multiple trajectory outputs and the trajectory with the highest probability is used for RMSE calculation. In other circumstances, our model generates a single trajectory, which is subsequently utilized for the evaluation.

\section{Experiment Results and Discussion}
In this section, we present our findings on the two datasets commonly used for trajectory prediction benchmarks. Four baseline models are compared based on the defined evaluation metrics in a three-lane highway scenario. Additionally, ablation studies are conducted to understand the significance of each component and try to provide deeper insights into the model design.

\subsection{Model Performance Comparison}
The compared results presented in Table \ref{tab:table2}. As can be found that, the proposed framework demonstrates good performance with respect to the RMSE across a prediction horizon of 50 frames when juxtaposed with existing baseline models. It exhibits a reduced loss in comparison to C-LSTM, CS-LSTM, and PiP. These outcomes suggest that the proposed model effectively captures salient features pertinent to long-term predictions. In summary, the proposed framework outperforms baseline models on the HighD dataset and delivers commendable performance on the NGSIM dataset.
 
\begin{table}[!h]
\caption{Prediction Error Obtained by Different Models in RMSE\label{tab:table2}}
\centering
\begin{tabular}{>{\centering\arraybackslash}m{0.8cm} >{\centering\arraybackslash}m{0.8cm} >{\centering\arraybackslash}m{0.8cm} >{\centering\arraybackslash}m{0.8cm} >{\centering\arraybackslash}m{0.8cm} >{\centering\arraybackslash}m{0.8cm} >{\centering\arraybackslash}m{0.8cm} >{\centering\arraybackslash}m{1cm}}
\hline\hline
Dataset & Horizon (Frame) & S-LSTM & CS-LSTM & PiP & GRIP & \textbf{GIMTP}\\
\hline
\multirow{5}{*}{NGSIM} & 10 & 0.65 &0.61 & 0.55 & 0.37 & \textbf{0.35} \\
 & 20 & 1.31 & 1.27 & 1.18 & 0.86 & \textbf{0.82} \\
 & 30 & 2.16 & 2.08 & 1.94 & 1.45 & \textbf{1.39} \\
 & 40 & 3.25 & 3.10 & 2.88 & \textbf{2.21} & 2.24 \\
 & 50 & 4.55 & 4.37 & 4.04 & 3.16 & \textbf{3.05} \\
\hline
\multirow{5}{*}{HighD} & 10 & 0.22 & 0.22 & 0.17 & 0.29 & \textbf{0.17} \\
 & 20 & 0.62 & 0.61 & 0.52 & 0.68 & \textbf{0.39} \\
 & 30 & 1.27 & 1.24 & 1.05 & 1.17 & \textbf{0.73} \\
 & 40 & 2.15 & 2.10 & 1.76 & 1.88 & \textbf{1.02} \\
 & 50 & 3.41 & 3.27 & 2.63 & 2.76 & \textbf{1.42} \\
\hline\hline
\end{tabular}
\end{table}

\begin{figure*}[!h]
\centering
\includegraphics[scale=0.38]{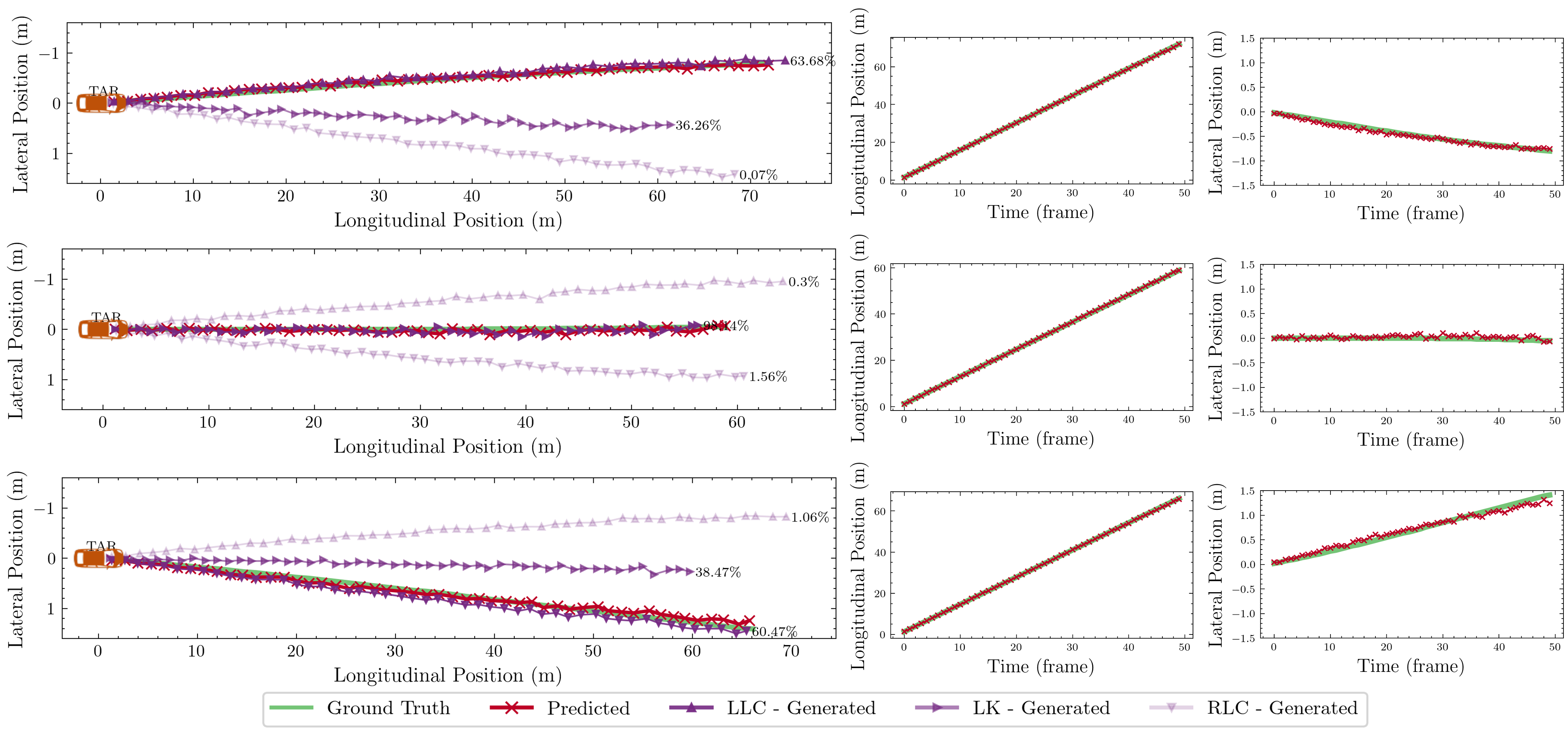}
\caption{Multi-modal trajectory generation results.}
\label{fig_9}
\end{figure*}

\begin{figure*}[!h]
\centering
\includegraphics[scale=0.5]{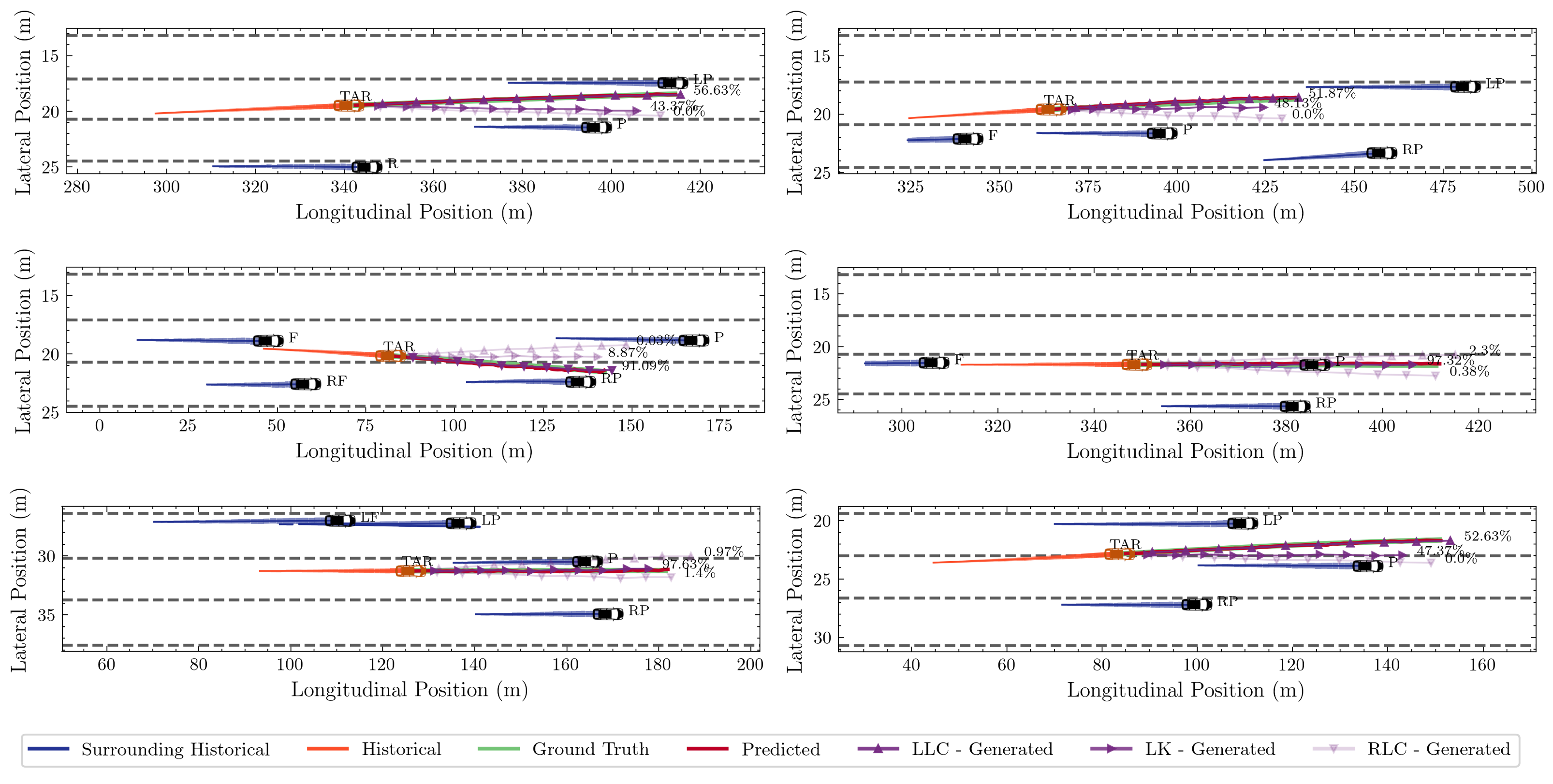}
\caption{Multi-modal trajectory prediction experiment results in three-lane highway scenarios.}
\label{fig_10}
\end{figure*}

\begin{figure*}[!h]
\centering
\includegraphics[scale=0.37]{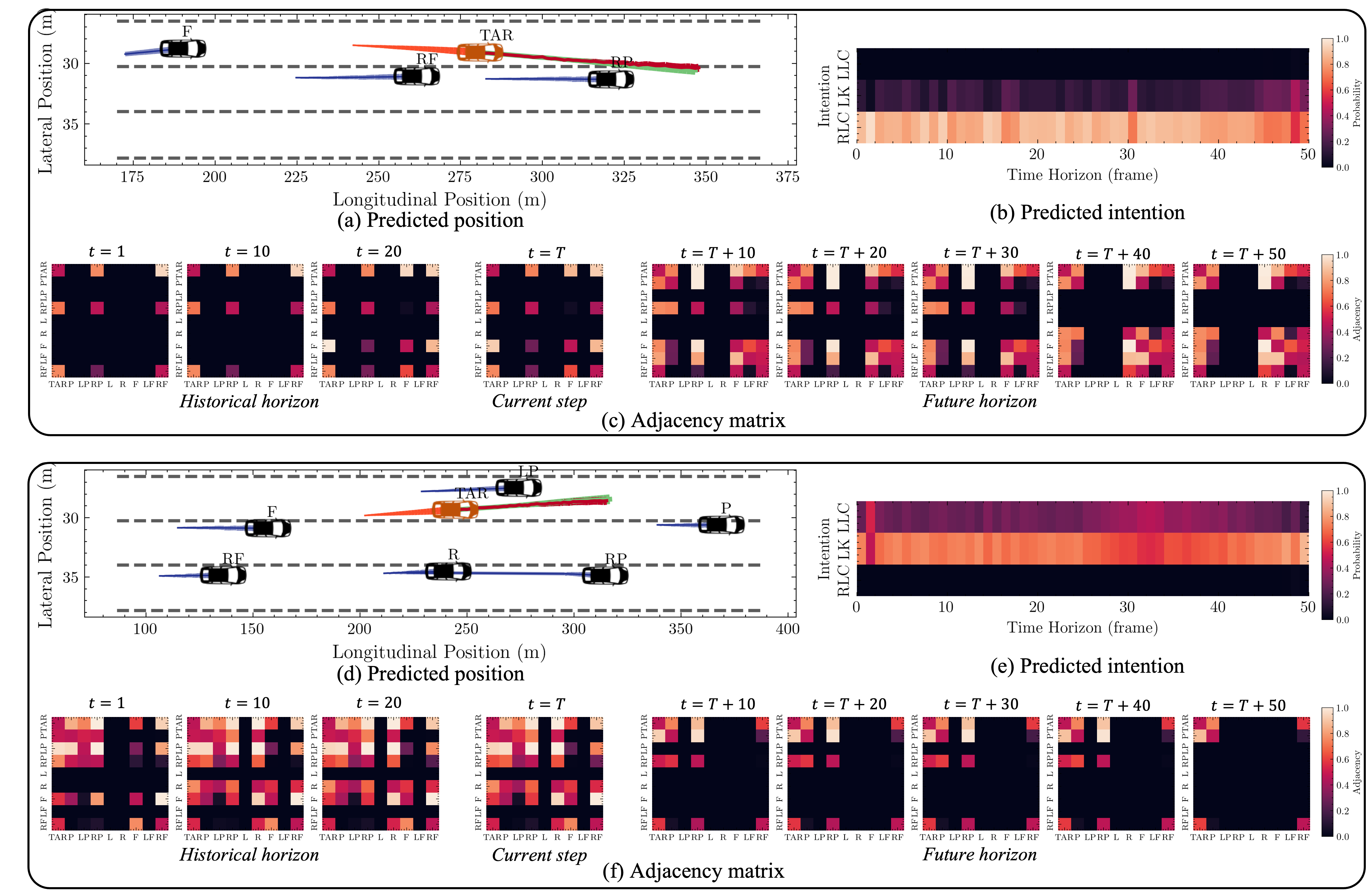}
\caption{Trajectory prediction experiment results. (a) and (d) Predicted trajectory v.s. ground truth trajectory. (b) and (e) Predicted lateral intention with probability. (c) and (f) Adjacency matrix of the vehicle group at historical and future time steps.}
\label{fig_11}
\end{figure*}

\subsection{Results of Multi-modal Predictions}
We begin by examining the performance of the multi-modal prediction. The results of experiments involving the prediction of multiple trajectories, considering different lateral intentions using the HighD dataset, are showcased in Figure \ref{fig_9}. The green line marks the ground truth trajectory, and the solid red line signifies the predicted trajectory using the fused intention features. Potential trajectories, contingent on the application of a lateral intention from the current time step, are depicted by the purple lines. In this experiment, the probabilities of LK, LLC, and RLC intentions are fixed at 1 throughout the entire future time horizon, enforcing mandatory lane changing or lane keeping for the predicted trajectories. The color opacity represents the probability of lane-changing intentions at the current time step $T$, with more solid colors indicating higher probabilities, and lighter colors representing lower probabilities. Figures \ref{fig_9}(b) and (c) depict the longitudinal and lateral positions over time, respectively. Figure \ref{fig_10} exhibits the prediction results of the target vehicles along with the historical trajectories of the surrounding vehicles across six experiments in three-lane highway scenarios, with the dashed grey lines denoting the lane markings' positions. Our model is proficient at predicting probabilities associated with each lateral and longitudinal intention, while concurrently forecasting the corresponding future trajectories for each intention. Moreover, the model allows for manual adjustments of the intentions and their respective probabilities at every time step within the future horizon $F$, which enables the observation of corresponding generated trajectories. This flexibility enriches the assessment of the model's performance across diverse scenarios. 

\subsection{Results of Trajectory, Intention, and Interaction Prediction}
Experiments to observe the predicted trajectories, intentions, and vehicle interactions are further conducted, depicted in Figure \ref{fig_11}. Figure \ref{fig_11}(b) and (e) illustrate the predicted intentions of the target vehicle at each time step within the future time horizon for the lane-changing scenario. Specifically, Figure \ref{fig_11}(b) delineates the right lane changing process, while Figure \ref{fig_11}(e) focuses on the lane keeping process of the target vehicle. Figure \ref{fig_11}(c) and (f) present the adjacency matrices of the vehicle group, generated by the historical graph embedding $\tilde H_T$ and the future-guided graph
embedding $\tilde H_F$ within the DGCN encoder of the Graph Interaction Encoder. These adjacency matrices represent vehicle interactions at the current time step, with lighter colors indicating more critical interactions. The model's generated future-guided adjacency matrix effectively predicts vehicles' entry into and exit from the vehicle group, along with their respective motion states. This future-guided adjacency matrix serves as a component of the input for Intention Predictor to generate fused hidden states in the future horizon. The proposed model successfully integrates precise future trajectory predictions with future intention estimations.

\subsection{Ablation Study}
Ablation Study is conducted to provide more insights into the performance of our model, especially the impact of different components on the prediction performance by disabling the corresponding component from the entire GIMTP. In particular, we consider the following three models, each of which removes a specific component.
\begin{itemize}
    \item \textbf{GIMTP w/o DGCN} variant use a simple GCN instead of the DGCN architecture in the Interaction Encoder. 
    \item \textbf{GIMTP w/o FG} variant does not provide a prediction of encoded future-guided graph matrix which denotes the future states in the Interaction Encoder and only uses historical motion states for intention prediction and feature fusion. 
    \item \textbf{GIMTP w/o FF} variant does not use feature fusion for each unique lateral and longitudinal intention and only uses a simple feature mapping from the hidden states generated from the Intention Encoder.
\end{itemize}
    
An investigation into the effects of model design variations, as presented in Table \ref{tab:table3}, reveals noteworthy implications. Upon eliminating various components from GIMTP, it is observed that the performance of the resulting model invariably experiences degradation to differing extents. Herein, certain observations of interest are elucidated. When the DGCN module is excluded, there is a distinct downturn in overall performance, suggesting that an increased number of DGCN layers is advantageous for dynamic graph-based information extraction from the low-level motion state of vehicles. Similarly, the omission of the FG module precipitates a substantial decrement in performance, which underscores the critical nature of the future-guided matrix embedding, generated within the Interaction Encoder. This suggests that the projected future motion states of the vehicle group can wield significant influence over the motion of the target vehicle. Lastly, the absence of the FF module underscores the vital role of feature fusion for each prospective behavioral intention. In summary, our ablation study revealed that the integration of the three distinct modules substantially enhances the model's performance metrics. Notably, the implementation of future-guided graph embedding and feature fusion exhibits a more significant impact compared to the use of the DGCN architecture. These findings underscore the relative importance of specific modular components in optimizing the overall effectiveness of the predictive model.

\begin{table}[!h]
\caption{Ablation Test Results in RMSE\label{tab:table3}}
\centering
\begin{tabular}{>{\centering\arraybackslash}m{1cm} >{\centering\arraybackslash}m{1cm} >{\centering\arraybackslash}m{1cm} >{\centering\arraybackslash}m{1cm} >{\centering\arraybackslash}m{1cm}}
\hline\hline
Horizon (Frame) & GIMTP w/o DGCN & GIMTP w/o FG & GIMTP w/o FF & GIMTP\\
\hline
10 & 0.17 & 0.20 & 0.19 & \textbf{0.17} \\
20 & 0.42 & 0.57 & 0.53 & \textbf{0.39} \\
30 & 0.79 & 1.14 & 0.98 & \textbf{0.73} \\
40 & 1.09 & 1.80 & 1.64 & \textbf{1.02} \\
50 & 1.62 & 2.27 & 2.33 & \textbf{1.42} \\
\hline\hline
\end{tabular}
\end{table}

\section{Conclusion}
In this study, we have introduced a Graph-based Interaction-aware Multi-modal Trajectory Prediction (GIMTP) framework that thoroughly investigates vehicle interactions, offering multiple potential predictions and estimating driving behavioral intentions in a probabilistic fashion. A dynamic adjacency matrix was constructed to comprehensively capture vehicle interactions, taking into account neighborhood, distance, and potential risk factors. The implementation of the DGCN structure enabled the encapsulation of both spatial and temporal vehicle interactions. The model embeds not only the historical motion state of the vehicle group but also the inferred future motion states, providing additional correlations and potential future interactions. The concept of feature fusion was employed to efficiently integrate historical and future embeddings for collective intention recognition and trajectory prediction, facilitating more precise trajectory generation based on latent variables symbolizing multi-modal behaviors. The model successfully anticipates both longitudinal and lateral driving behaviors in a multi-modal fashion, associating potential future trajectories with corresponding probabilities.

Nevertheless, the graph modeling necessitates a fixed configuration of the vehicle group to estimate the target vehicle's motion states. The motion states of each vehicle within the surrounding environment could also be influenced by their respective surrounding vehicles, which is a factor that is not fully explored in this study. Additionally, constructing a dynamic graph for a long sequence is not computationally efficient. In subsequent work, our aim is to improve the vehicle group's adaptability and further investigate more flexible formulations of the vehicle group, with the purpose of more effectively identifying and embedding interactions. We also propose the incorporation of traffic flow theory to discern mesoscopic motion patterns within the traffic environment and to compress these patterns accordingly.


\begin{IEEEbiography}[{\includegraphics[width=1in,height=1.25in,clip,keepaspectratio]{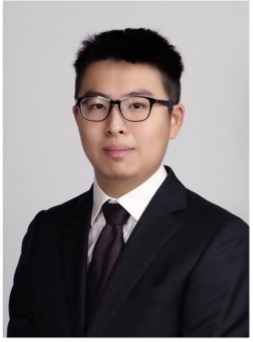}}]{Keshu Wu}
received his B.S. degree in Civil Engineering from Southeast University, Nanjing, China, in 2017. He received a M.S. degree in Civil and Environmental Engineering from Carnegie Mellon University in 2018, and a M.S. degree in Computer Sciences from the University of Wisconsin-Madison in 2022. He is currently a Ph.D. student in Civil and Environmental Engineering at the University of Wisconsin-Madison. His research interests include the application and innovation of machine learning and deep learning techniques in the transportation system and connected automated vehicles highway.
\end{IEEEbiography}

\begin{IEEEbiography}[{\includegraphics[width=1in,height=1.25in,clip,keepaspectratio]{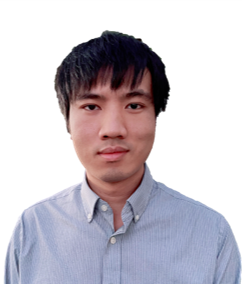}}]{Haotian Shi}
serves as a research associate at University of Wisconsin-Madison. He received his Ph.D. degree in Civil and Environmental Engineering from the University of Wisconsin-Madison in May 2023. He also received three M.S. degrees in Power and Machinery Engineering (Tianjin University, 2020), Civil and Environmental Engineering (UW-Madison, 2020), and Computer Sciences (UW-Madison, 2022). His main research directions are prediction/control of connected automated vehicles, intelligent transportation systems, traffic crash data analysis, and deep reinforcement learning.
\end{IEEEbiography}

\begin{IEEEbiography}[{\includegraphics[width=1in,height=1.25in,clip,keepaspectratio]{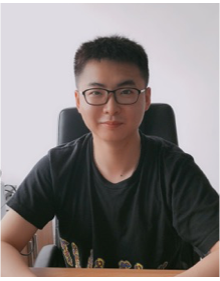}}]{Yang Zhou}
received the Ph.D. degree in Civil and Environmental Engineering from University of Wisconsin Madison, WI, USA, in 2019, and the M.S. degree in Civil and Environmental Engineering from  University of Illinois at Urbana-Champaign, Champaign, IL, USA, in 2015. He is currently an assistant professor in the Zachry Department of Civil and Environmental Engineering, and Career Initiation Fellow in the Institute of Data Science, Texas A\&M University. Before joining Texas A\&M, he is a postdoctoral researcher in civil engineering, University of Wisconsin Madison, WI, USA. He is current an member in TRB traffic flow theory CAV subcommittee, network modeling CAV subcommittee and American Society of Civil Engineering TDI-AI committee. His main research directions are connected automated vehicles robust control, interconnected system stability analysis, traffic big data analysis, and microscopic traffic flow modeling.
\end{IEEEbiography}

\begin{IEEEbiography}[{\includegraphics[width=1in,height=1.25in,clip,keepaspectratio]{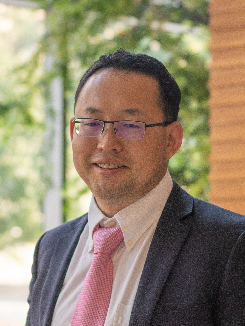}}]{Xiaopeng (Shaw) Li}
received the B.S. degree in civil engineering with a computer engineering minor from Tsinghua University, China, in 2006, and the M.S. degree in civil engineering, the M.S. degree in applied mathematics, and the Ph.D. degree in civil engineering from the University of Illinois at Urbana-Champaign, USA, in 2007, 2010, and 2011, respectively. He is currently an Associate Professor with the Department of CEE, USF. He is the Director of the National Institute for Congestion Reduction (NICR). His major research interests include automated vehicle traffic control and connected and interdependent infrastructure systems. He was a recipient of the National Science Foundation CAREER Award.
\end{IEEEbiography}

\begin{IEEEbiography}[{\includegraphics[width=1in,height=1.25in,clip,keepaspectratio]{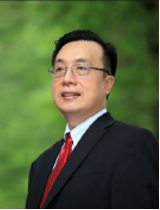}}]{Bin Ran}
is the Vilas Distinguished Achievement Professor and Director of ITS Program at the University of Wisconsin at Madison. Dr. Ran is an expert in dynamic transportation network models, traffic simulation and control, traffic information system, Internet of Mobility, Connected Automated Vehicle Highway (CAVH) System. He has led the development and deployment of various traffic information systems and the demonstration of CAVH systems. Dr. Ran is the author of two leading textbooks on dynamic traffic networks. He has co-authored more than 240 journal papers and more than 260 referenced papers at national and international conferences. He holds more than 20 patents of CAVH in the US and other countries. He is an associate editor of Journal of Intelligent Transportation Systems.
\end{IEEEbiography}
\vspace{11pt}
\vfill

\end{document}